\documentclass[preprint,12pt,authoryear]{elsarticle}
\usepackage[T1]{fontenc}
\usepackage{amsmath,amssymb,bm}
\usepackage{graphicx,booktabs,tabularx,array}
\usepackage{xspace}
\usepackage[hidelinks]{hyperref}

\newcommand{\model}{MEOX\xspace}
\newcommand{\R}{\mathbb{R}}
\newcolumntype{Y}{>{\raggedright\arraybackslash}X}
\graphicspath{{figures/}}
\hypersetup{pdftitle={MEOX: Modality-Conditioned Routing Before Fusion in a Compact Earth Observation Encoder},pdfauthor={Mohanad Albughdadi}}
\journal{ISPRS J. Photogramm. Remote Sens.}

\begin{document}
\begin{frontmatter}

\title{MEOX: Modality-Conditioned Routing Before Fusion in a Compact Earth Observation Encoder}

\author[ecmwf]{Mohanad Albughdadi\corref{cor1}}
\ead{mohanad.albughdadi@ecmwf.int}
\cortext[cor1]{Corresponding author}
\affiliation[ecmwf]{organization={European Centre for Medium-Range Weather Forecasts (ECMWF)},
                    city={Bonn},
                    postcode={53175},
                    country={Germany}}

\begin{abstract}
Multimodal Earth Observation models often fuse radar and optical inputs early, limiting sensor-conditioned processing, or maintain separate sensor streams through the encoder, increasing computational cost. We present MEOX (Multimodal Earth Observation with eXperts), which delays fusion until after a shared mixture-of-experts block that processes Sentinel-2 and ascending and descending Sentinel-1 streams independently. Co-located patches are then fused with validity-aware weights into a single sequence processed by fourteen shared blocks, reducing the sequence length from \(3N+5\) to \(N+5\) tokens for \(N\) patches. We pretrain the 2.939M-parameter encoder on 1.228M MMEarth64 samples using masked radar-optical reconstruction. Under the reported frozen-transfer protocols, MEOX reaches 90.56\% average class accuracy on EuroSAT and 64.42\% mean intersection over union (mIoU) on cashew segmentation, 2.26 and 5.02 points above the strongest reported cross-sensor mixture-of-experts (CSMoE) variants. Full-encoder adaptation increases BigEarthNet micro-average precision from 57.22\% to 72.95\%. In a controlled same-corpus test using synthetic random patch removal, masking 75\% of Sentinel-2 patches reduces WorldCover mIoU from 30.45\% to 26.53\% when both Sentinel-1 streams remain available; with Sentinel-2 alone, mIoU falls to 13.81\%. For identical zeroed optical patches, declaring them missing through the validity mechanism improves mIoU from 19.42\% to 26.53\%. These results characterize a compact route-then-fuse encoder that retains sensor-conditioned processing and uses complementary radar observations when optical inputs are incomplete. The model and experiments code are available on \url{https://github.com/AlbughdadiM/compact-multimodal-moe-eo}.
\end{abstract}

\begin{keyword}
Earth observation \sep multimodal fusion \sep mixture of experts \sep Sentinel-1 \sep Sentinel-2 \sep compact representation learning
\end{keyword}

\end{frontmatter}

\section{Introduction}
Multimodal Earth Observation (EO) representation learning exploits complementary optical, synthetic aperture radar (SAR), and environmental observations, and must remain usable when some inputs are unavailable. A central architectural choice in these models is when the sensor streams should interact. Retaining separate token streams inside a common backbone increases sequence length and attention cost. Fusing immediately after patch projection reduces these costs, but removes the opportunity for sensor-conditioned attention and routing before the observations are combined.

Recent EO models emphasize broad modality support, multiscale imagery, and large pretraining corpora \citep{nedungadi2024mmearth,fuller2023croma,xiong2024dofa,astruc2025anysat,jakubik2025terramind}. This work investigates a narrower but important challenge. It studies whether an encoder can preserve sensor-conditioned computation without carrying modality multiplicity through its full depth. This question matters when processing large archives or deploying models under storage and memory constraints. Sparse mixture-of-experts (MoE) layers provide conditional computation, but conventional experts can increase storage, and balanced routing alone does not establish meaningful specialization~\citep{pan2026risesparsemixtureofexpertssurvey}.

MEOX addresses this trade-off through a \emph{route-then-fuse} factorization. A single shared transformer-MoE block processes each available Sentinel-1~(S1) and Sentinel-2~(S2) stream independently, allowing attention and expert routing to respond separately to each stream. Learned validity-aware weights then fuse corresponding spatial patches, and the remaining fourteen blocks process one common sequence. The dominant encoder depth is therefore independent of the number of supplied streams. Named-band adapters and explicit missingness signals provide a consistent interface for configured sensor and band subsets. The compact expert parameterization uses staged hidden widths and projections shared among experts within each layer \citep{albughdadi2025geomoemae}; the multimodal contribution is the placement of shared sparse processing before patchwise fusion.

The contributions are:
\begin{enumerate}
\item A route-then-fuse multimodal factorization that applies shared-weight, sensor-conditioned sparse processing before reducing all available rasters to one spatial sequence. With three streams, the deeper encoder processes $N+5$ rather than $3N+5$ tokens.
\item A compact realization in which named-band adapters, explicit validity signals, shared expert projections, and low-rank expert residuals support configured input subsets while limiting the encoder to 2.939M parameters.
\item An empirical characterization of the resulting resource--accuracy operating point through six frozen transfer tasks, two spatial protocols, BigEarthNet adaptation, retrieval, metadata ablation, controlled sensor removal and degradation, and direct measurements of routing activity, sensor association, spatial dependence, and expert contribution.
\end{enumerate}

\section{Related Work}
\subsection{Masked and Multimodal EO Representation Learning}

Masked autoencoders learn representations by reconstructing hidden image content with an asymmetric encoder--decoder architecture \citep{he2022mae}. SatMAE applies masked modeling to temporal and multispectral satellite imagery \citep{cong2022satmae}, while Scale-MAE accounts for physical image scale \citep{reed2023scalemae}. MultiMAE extends masked reconstruction across modalities and prediction targets \citep{bachmann2022multimae}. MMEarth studies multimodal geospatial pretraining on a globally distributed archive and provides the pretraining data used in this work \citep{nedungadi2024mmearth}.

CROMA combines masked reconstruction with radar--optical contrastive learning \citep{fuller2023croma}. Its contrastive objective explicitly aligns radar and optical embeddings in a common latent space. Reconstruction through modality-specific output heads does not impose the same alignment, even when one encoder processes both sensors.

Several EO foundation models broaden the input interface or the range of supported observations. DOFA conditions image encoding on spectral wavelength \citep{xiong2024dofa}, and AnySat supports multiple modalities, resolutions, and spatial extents \citep{astruc2025anysat}. TerraMind studies large-scale generative multimodal pretraining \citep{jakubik2025terramind}. Prithvi-EO-2.0 incorporates temporal and location information \citep{szwarcman2024prithvi2}, while Presto uses a lightweight architecture for multispectral time series and metadata \citep{tseng2023presto}.

Other models introduce mechanisms for particular multimodal settings. DAMoE conditions sparse routing on frequency, spatial, and spectral priors for EO fusion and reconstruction \citep{liu2026damoe}. SimpFit handles missing target-domain modalities through learned proxy tokens and lightweight affine adaptation \citep{xu2026simpfit}. HATFormer models interactions between optical imagery and height information for semantic and height change detection \citep{liu2025hatformer}. Mind the Modality Gap maps additional remote-sensing modalities into CLIP's representation space \citep{zavras2025modalitygap}, and SARCLIP learns SAR--text representations from more than 400,000 paired samples \citep{jiang2026sarclip}.

MEOX examines fusion placement in a compact spatial S1/S2 encoder. One shared sparse block processes each available sensor stream independently before validity-aware weights combine co-located patches. The remaining encoder layers receive only the fused spatial sequence. This arrangement retains sensor-conditioned attention and routing before fusion while keeping the deeper token count independent of the number of supplied sensor streams.

\subsection{Experts, Parameter Sharing, and Spatial Encoding}
Sparse top-$k$ routing originates in large MoE systems \citep{shazeer2017moe} and has been applied to vision transformers \citep{riquelme2021vmoe}. Switch Transformers simplify balancing through a probability assignment product \citep{fedus2022switch}. Soft MoE instead aggregates tokens into expert slots \citep{puigcerver2024softmoe}; CSMoE applies this approach to cross-sensor EO learning and supplies the closest published benchmark comparison \citep{hackel2026csmoe}. MEOX does not introduce a new routing rule. It uses discrete top-2 routing as the conditional operator in the shared pre-fusion stage and throughout the compact backbone, with no capacity pruning or forced dispatch quotas. The experiments therefore measure executed routes and expert interventions rather than treating the presence of an MoE layer as evidence of specialization.

Geo-MoE-MAE introduced the compact staged-expert backbone, decreasing expert hidden widths, and value/output projections shared among experts within each layer \citep{albughdadi2025geomoemae}. MEOX uses this parameterization and adds expert-private rank-8 residual projections, related to low-rank adaptation \citep{hu2022lora}. Rotary positional embeddings \citep{su2024roformer,heo2024ropevit} replace an additive absolute-position table. They permit variable token grids but do not establish invariance to physical ground-sampling distance. Table~\ref{tab:changes} separates the retained backbone choices from the multimodal design introduced here.

\begin{table}[!h]
\caption{Architectural relationship between Geo-MoE-MAE and MEOX.}
\label{tab:changes}
\centering\footnotesize
\begin{tabularx}{\columnwidth}{@{}p{0.19\columnwidth}YY@{}}
\toprule
Component & Geo-MoE-MAE & \model \\
\midrule
Raster inputs & Seven-band Landsat & S2, S1 ascending/descending; named subsets \\
Metadata & Lat., lon., week, hour & Lat., lon., month, ECMWF Reanalysis v5; explicit missingness \\
Fusion & Single raster stream & Shared stream block, then validity-aware gating \\
Position & Additive learned table & Axial rotary embeddings in attention \\
Experts & Shared value/output & Shared value/output + rank-8 residuals \\
Decoder & Two self-attention blocks with top-2 MoE feed-forward layers
        & Two self-attention blocks with dense SwiGLU feed-forward layers \\
Objective & Masked-patch mean squared error (MSE) + weighted visible-patch MSE + summed coefficient-of-variation load and importance penalties
          & Stream-averaged masked MSE over valid target elements + depth-averaged Switch-style routing balance \\
\bottomrule
\end{tabularx}
\end{table}

\section{Method}
\label{sec:method}
\subsection{Inputs and Missingness}
We use \emph{sensor stream} for each configured raster input: S2, ascending S1 (S1A), or descending S1 (S1D). S1 and S2 are the two sensing modalities, while S1A and S1D are separate acquisition streams. Let $\mathcal M$ be the configured stream set and $\mathcal A\subseteq\mathcal M$ the streams supplied to the encoder. All supplied rasters are aligned to a common $H\times W$ grid before tokenization. For stream $m$, $X_m\in\R^{B\times C_m\times H\times W}$ contains a batch of $B$ rasters with $C_m$ configured channels, and $V_m\in\{0,1\}^{B\times C_m\times H\times W}$ marks valid elements. Band names map supplied channels into the configured physical order. The vector $r_m\in\{0,1\}^{C_m}$ marks which configured bands are present. Missing elements are zero-filled only after $V_m$ has retained their missing state, so a missing value remains distinguishable from a valid observation equal to its normalization mean.

Each sensor image is divided into nonoverlapping $P\times P$ patches. An image of size $H\times W$ therefore produces $N=HW/P^2$ spatial tokens. For sensor $m$, the token representing patch $i$ is
\begin{equation}
t_{m,i}=A_m(X_m)_i+Q_m(V_m)_i+\epsilon_m+R_m r_m.
\label{eq:adapter}
\end{equation}
In Eq.~\eqref{eq:adapter}, $A_m$ and $Q_m$ map a $C_m\times H\times W$ raster or validity mask to $N$ vectors in $\R^d$. The term $A_m(X_m)_i$ is the patch-content embedding, and $Q_m(V_m)_i$ embeds the corresponding validity mask. The validity projection is initialized to zero so that it does not perturb the initial image representation. The learned vector $\epsilon_m\in\R^d$ identifies the stream, and $R_m\in\R^{d\times C_m}$ projects the band-presence vector. The four width-$d$ terms form one stream-conditioned, validity-aware patch token. In the active configuration, S2 has 13 bands, while S1A and S1D each use VV and VH\@. With $H=W=64$ and $P=4$, every stream produces $N=16\times16=256$ tokens of width $d=144$.

\begin{figure}[!h]
\centering
\includegraphics[width=\textwidth]{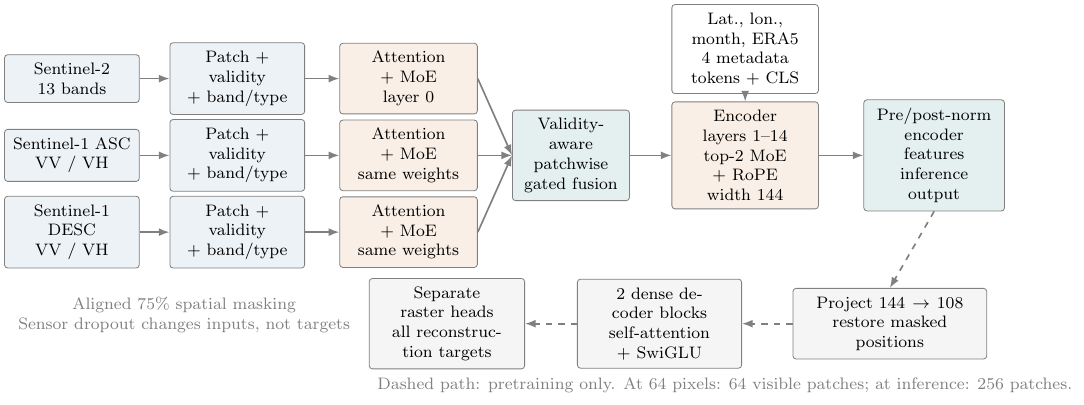}
\caption{\model pretraining. All supplied rasters share spatial mask indices. The same layer-0 parameters process each stream independently; gated fusion yields one spatial sequence. Metadata and the classification (CLS) token enter only after fusion. The dense decoder reconstructs every target stream, including streams deliberately removed from encoder inputs. At inference the decoder and spatial mask are omitted, and pre- or post-norm encoder features are returned.}
\label{fig:architecture}
\end{figure}

Metadata is represented by four additional tokens, one each for latitude, longitude, month, and European Centre for Medium-Range Weather Forecasts Reanalysis v5 (ERA5) climate context. Latitude, longitude, and month are represented by two sine/cosine components, while ERA5 provides 12 temperature and precipitation variables covering recent and longer time windows. For one sample and a metadata group $j$ with $D_j$ components, let $x_j\in\R^{D_j}$ contain its values and $v_j\in\{0,1\}^{D_j}$ indicate which components are available. Its token is
\begin{equation}
u_j=W_j(x_j\odot v_j)+\gamma_j+\epsilon_j^{\rm type}
+(1-\overline v_j)\epsilon_j^{\rm miss}.
\label{eq:metadata}
\end{equation}
where $W_j\in\R^{d\times D_j}$, $\gamma_j\in\R^d$ is its bias, $\odot$ denotes elementwise multiplication, and $\overline v_j=D_j^{-1}\sum_{k=1}^{D_j}v_{j,k}$ is the available-component fraction. In Eq.~\eqref{eq:metadata}, the first term projects available values, $\epsilon_j^{\mathrm{type}}\in\R^d$ identifies the metadata group, and $\epsilon_j^{\mathrm{miss}}\in\R^d$ represents missingness. The missing-state contribution is zero when all components are present and increases as components become unavailable. Omitting a metadata group therefore produces an explicit missing-data representation instead of a genuine value of zero. Figure~\ref{fig:architecture} summarizes the complete pretraining and inference paths.

\subsection{Route-Then-Fuse Sensor Processing}
The route-then-fuse stage preserves sensor-conditioned computation without allocating a complete backbone to each stream. Before fusion, MEOX processes each stream $m\in\mathcal A$ independently using the same transformer-MoE block. Let $t_m$ be its patch-token sequence. The shared block produces
\[
z_m=F_\theta(t_m),
\]
where $F_\theta$ denotes the shared block with parameters $\theta$, and $z_{m,i}\in\mathbb{R}^d$ is its representation of spatial patch $i$. Although $\theta$ is shared, every stream sequence is processed in a separate forward pass. Attention and expert routing therefore respond to stream-conditioned observations before cross-stream fusion.

At each spatial position, the model computes one fusion score for every available stream:
\begin{align}
a_{m,i}
&=
w_f^\top\operatorname{LN}(z_{m,i})
+c_m
+\log\!\left(\max(q_{m,i},10^{-6})\right), \\
\alpha_{m,i}
&=
\frac{\exp(a_{m,i})}
{\sum_{m'\in\mathcal A}\exp(a_{m',i})},
\qquad
z_i
=
\sum_{m\in\mathcal A}\alpha_{m,i}z_{m,i}.
\label{eq:fusion}
\end{align}
Here, $\operatorname{LN}$ is LayerNorm, $a_{m,i}$ is the unnormalized fusion score, $w_f\in\mathbb{R}^d$ is a learned scoring vector, and $c_m$ is a learned stream-specific bias. The quantity $q_{m,i}\in[0,1]$ is the fraction of valid elements in patch $i$ of stream $m$. Its logarithm acts as a validity prior: it is zero for a fully valid patch and increasingly negative as the valid fraction decreases. The lower bound $10^{-6}$ prevents numerical evaluation of $\log(0)$.
The softmax in Eq.~\eqref{eq:fusion} produces nonnegative fusion weights $\alpha_{m,i}$ that sum to one across the available streams. Their weighted sum gives the fused patch representation $z_i$. Streams that are not supplied are excluded from both the softmax and the weighted sum. If all supplied streams are invalid at a position, the lower bound keeps the computation finite, while the corresponding invalid target elements remain excluded from the reconstruction loss. The fusion parameters $w_f$ and $c_m$ are initialized to zero, so equally valid streams initially receive equal weights. Because the scores are computed separately at each patch, the model can later learn to favor different streams at different spatial positions.

After fusion, the four metadata tokens and one learned classification (CLS) token are prepended to the fused patch sequence. Fourteen additional transformer-MoE blocks then process this common sequence. With three streams, the shared pre-fusion block is evaluated three times, but the deeper encoder processes only one spatial stream. Let $N_{\mathrm{vis}}$ be the number of spatial patches retained after masking. After fusion, the deeper encoder processes \(N_{\mathrm{vis}}+5\) tokens during masked pretraining and \(N+5\) tokens during inference. Concatenating all three stream sequences instead would increase these lengths to \(3N_{\mathrm{vis}}+5\) and \(3N+5\). We call the residual sequence immediately before the encoder's final LayerNorm \emph{pre-norm} features and the normalized output \emph{post-norm} features.

\subsection{Compact Sparse Experts}
Each encoder block contains a pre-normalized self-attention sublayer followed by a pre-normalized MoE feed-forward sublayer, both with residual connections. Layer $l$ contains $E_l$ experts: three in layers 0--4, four in layers 5--9, and five in layers 10--14. The expert hidden width decreases gradually from 144 in the first layer to 72 in the last layer, while the token width remains $d=144$ throughout the encoder. All encoder blocks use eight attention heads.
Let $x\in\mathbb{R}^d$ be one normalized token entering the MoE sublayer. To reduce clutter, the layer index $l$ is omitted from router logits, gates, and projection matrices below. These parameters are independent across layers; within a layer, only matrices explicitly identified below are shared across experts. The router computes one clean logit for each of the $E_l$ experts:
\[
\ell(x)=W_r x+b_r,
\]
where $W_r\in\mathbb{R}^{E_l\times d}$ and $b_r\in\mathbb{R}^{E_l}$ are learned router parameters. It also computes a positive, input-dependent noise scale:
\[
s(x)=\operatorname{softplus}(W_s x+b_s)+\delta,
\]
where $W_s\in\mathbb{R}^{E_l\times d}$ and $b_s\in\mathbb{R}^{E_l}$ are learned parameters, and $\delta=10^{-9}$ prevents a zero noise scale. During training, the routing logits are
\[
\tilde{\ell}(x)=\ell(x)+\eta\odot s(x),
\qquad
\eta\sim\mathcal N(0,I_{E_l}),
\]
with independent Gaussian samples for each token and expert. During evaluation, routing is deterministic and uses
\[
\tilde{\ell}(x)=\ell(x).
\]

Let
\[
\mathcal K(x)=\operatorname{Top2}\!\left(\tilde{\ell}(x)\right)
\]
denote the indices of the two experts with the largest routing logits. The weight assigned to expert $e$ is
\begin{equation*}
g_e(x)
=
\mathbf{1}\!\left[e\in\mathcal K(x)\right]
\frac{\exp\!\left(\tilde{\ell}_e(x)\right)}
{\sum_{e'\in\mathcal K(x)}
 \exp\!\left(\tilde{\ell}_{e'}(x)\right)}.
\end{equation*}
Here, $\mathbf{1}[e\in\mathcal K(x)]$ is one if expert $e$ is selected and zero otherwise. Thus, unselected experts receive zero weight, while the weights of the two selected experts are positive and sum to one. The MoE output is
\begin{equation*}
\operatorname{MoE}_l(x)
=
\sum_{e=1}^{E_l}g_e(x)\mathcal E_{l,e}(x),
\end{equation*}
where $\mathcal E_{l,e}(x)$ is the transformation produced by expert $e$ in layer $l$. Both selected experts are evaluated. Because MEOX does not impose a per-expert capacity limit, no token is dropped when an expert receives many assignments.
Each expert is a SwiGLU feed-forward network \citep{shazeer2020glu}. Let $d_h$ denote the hidden width of the current layer and let $\rho=8$ be the rank of the expert-specific residual paths. Omitting biases and output dropout for readability, expert $e$ computes
\begin{align*}
h_e
&=
\operatorname{SiLU}(W_{g,e}x)
\odot
\left(
W_v x+U_{v,e}D_{v,e}x
\right), \\
\mathcal E_{l,e}(x)
&=
W_o h_e+U_{o,e}D_{o,e}h_e.
\end{align*}
The matrix
\[
W_{g,e}\in\mathbb{R}^{d_h\times d}
\]
is the expert-specific SwiGLU gating projection. The value and output projections
\[
W_v\in\mathbb{R}^{d_h\times d},
\qquad
W_o\in\mathbb{R}^{d\times d_h}
\]
are shared by all experts within the same layer.

Each expert also has two private rank-$\rho$ residual paths. Their dimensions are
\[
D_{v,e}\in\mathbb{R}^{\rho\times d},
\qquad
U_{v,e}\in\mathbb{R}^{d_h\times\rho}
\]
for the value projection, and
\[
D_{o,e}\in\mathbb{R}^{\rho\times d_h},
\qquad
U_{o,e}\in\mathbb{R}^{d\times\rho}
\]
for the output projection. These low-rank paths allow each expert to modify the shared transformations without storing complete private copies of $W_v$ and $W_o$. The $U$ matrices are initialized to zero, so the private residual contributions initially vanish and emerge during training. The projections are shared only among experts in the same layer; different encoder layers have independent parameters.

\subsection{Position and Reconstruction Decoder}
Axial two-dimensional rotary positional embeddings (RoPE) rotate attention queries and keys using the runtime patch coordinates. No absolute-position vector is added to the residual stream. This removes a direct positional input to the router, although contextual features may still contain spatial information. Both image dimensions must be divisible by $P$; rectangular inputs are supported.

During pretraining, 75\% of spatial patches are masked using the same indices across streams. The encoder sees 64 visible patches at 64 pixels. Its post-norm outputs are projected to width 108, visible patch positions are restored, and a learned mask token fills the remaining locations. The five prefix tokens are retained. Two dense transformer blocks with RoPE, SwiGLU hidden width 54, and separate stream heads predict all patch elements. The decoder has 176,096 parameters, or 5.65\% of the 3,114,993-parameter complete model. It has neither MoE layers nor cross-attention shortcuts.

\subsection{Pretraining Objective}
During pretraining, stream dropout changes only the inputs given to the encoder. The reconstruction targets are created before dropout and therefore still contain all three streams: S2, S1A, and S1D. In 90\% of batches, all three streams are supplied to the encoder. In the remaining 10\%, one of the following six input combinations is selected uniformly:
\[
\begin{aligned}
&\{\mathrm{S2}\},\;
\{\mathrm{S1}_{\mathrm{asc}}\},\;
\{\mathrm{S1}_{\mathrm{desc}}\},\;
\{\mathrm{S2},\mathrm{S1}_{\mathrm{asc}}\},\\
&\{\mathrm{S2},\mathrm{S1}_{\mathrm{desc}}\},\;
\{\mathrm{S1}_{\mathrm{asc}},\mathrm{S1}_{\mathrm{desc}}\}.
\end{aligned}
\]
The decoder must still reconstruct all three targets, including streams omitted from the encoder input. This trains the model to operate with missing streams and to predict one stream from the others. Each stream is absent in three of the six dropout combinations and is therefore omitted in approximately 5\% of training batches. The same stream combination is used for every sample in a batch. Individual bands are not randomly removed in this pretraining run.
The reconstruction loss is calculated only on spatially masked and valid target values. Let $m$ identify a stream, $n$ a sample in the batch, $i$ a spatial patch, and $k$ an element within that patch, including its pixel and channel position. Let $Y_{m,n,i,k}$ be the target value and $\widehat{Y}_{m,n,i,k}$ the corresponding decoder prediction. The binary mask
\[
M_{n,i}=
\begin{cases}
1, & \text{if patch }i\text{ is masked},\\
0, & \text{if patch }i\text{ is visible}
\end{cases}
\]
is shared across sensors. The target-validity indicator, obtained by patchifying and indexing the corresponding raster mask $V_m$, is
\[
\Omega_{m,n,i,k}=
\begin{cases}
1, & \text{if the target element is valid},\\
0, & \text{if it is nodata or unavailable}
\end{cases}.
\]
This indicator excludes invalid values. The reconstruction loss for sensor $m$ is
\begin{equation*}
\mathcal L_m
=
\frac{
\sum_{n,i,k}
M_{n,i}\Omega_{m,n,i,k}
\left(
\widehat{Y}_{m,n,i,k}-Y_{m,n,i,k}
\right)^2
}{
\max\!\left(
1,\,
\sum_{n,i,k}M_{n,i}\Omega_{m,n,i,k}
\right)
}.
\end{equation*}
The numerator is the total squared error over valid elements in masked patches. The denominator is the number of such elements, so $\mathcal L_m$ is their mean squared error. The maximum with one prevents division by zero when a sensor has no valid masked target in a batch. Visible patches and invalid target values do not contribute to this loss.
Let $\mathcal M_{\mathrm{valid}}\subseteq\mathcal M$ be the set of streams having at least one valid masked target in the current batch. Their losses are averaged:
\begin{equation*}
\mathcal L_{\mathrm{rec}}
=
\frac{1}{|\mathcal M_{\mathrm{valid}}|}
\sum_{m\in\mathcal M_{\mathrm{valid}}}\mathcal L_m.
\end{equation*}
Each stream therefore contributes one stream-level loss, regardless of its number of channels. Without this normalization, the 13-band S2 target would contribute approximately $13/2=6.5$ times as many error terms as either two-band S1 target. Equal stream averaging prevents this channel-count difference from dominating training.
The final objective also includes an auxiliary term that encourages balanced expert routing. In encoder layer $l$, let $T_l$ be the number of routed tokens. For token $x_t$, let $p_{l,e}(x_t)$ be the clean softmax probability assigned to expert $e$, before top-2 selection. Its average over tokens is
\begin{equation*}
P_{l,e}
=
\frac{1}{T_l}
\sum_{t=1}^{T_l}p_{l,e}(x_t).
\end{equation*}
Let $\mathcal K_l(x_t)$ contain the two experts executed for token $x_t$. The normalized fraction of executed assignments received by expert $e$ is
\begin{equation*}
f_{l,e}
=
\frac{1}{2T_l}
\sum_{t=1}^{T_l}
\mathbf{1}\!\left[e\in\mathcal K_l(x_t)\right].
\end{equation*}
Because every token selects two experts, both $\sum_e P_{l,e}$ and $\sum_e f_{l,e}$ equal one. The executed fractions $f_{l,e}$ are detached from gradient computation because top-2 selection is discrete. The balance loss for layer $l$ is
\begin{equation*}
\mathcal B_l
=
E_l\sum_{e=1}^{E_l}f_{l,e}P_{l,e}.
\end{equation*}
This term is approximately one under uniform routing and becomes larger when router probabilities and executed assignments concentrate on a small number of experts.
Let $L_{\mathrm{enc}}=15$ be the number of encoder layers. The complete pretraining objective is
\begin{equation}
\mathcal L_{\mathrm{total}}
=
\mathcal L_{\mathrm{rec}}
+
\frac{\beta}{L_{\mathrm{enc}}}
\sum_{l=0}^{L_{\mathrm{enc}}-1}\mathcal B_l,
\qquad
\beta=0.01.
\label{eq:loss}
\end{equation}
Equation~\eqref{eq:loss} averages the balance term over depth so that its scale does not increase with the number of encoder layers. The shared pre-fusion block, layer 0, is evaluated separately for every supplied stream, and its per-stream balance losses are first averaged into one value $\mathcal B_0$. No position penalty, dispatch guard, latent alignment objective, visible-patch reconstruction loss, or auxiliary semantic loss is included.
\section{Experimental Protocol}
\subsection{MMEarth64 Pretraining}
The MMEarth64 release used here contains 1,240,526 samples \citep{nedungadi2024mmearth}. Because the release assigns every sample to its training split, we permute those indices with NumPy random seed 42 and reserve the first 1\% as validation. This deterministic fallback yields 1,228,121 training and 12,405 validation samples. We use S2, S1A/S1D VV/VH, and the four metadata groups. Dynamic World and WorldCover labels are not pretraining targets. Release-provided band statistics normalize valid rasters; S2 statistics follow the tile's L1C/L2A metadata, and S1 uses the appropriate orbit-channel offsets. Nodata sentinels and nonfinite values are identified before normalization.

Training uses AdamW for 50 epochs, batch size 128, peak learning rate $3\times10^{-4}$, weight decay 0.05, and gradient clipping at 1.0. A linear warmup covers 5\% of optimizer steps, followed by cosine decay. Biases, normalization parameters, and learned token/type/missingness embeddings are excluded from weight decay. Mixed precision is used on an NVIDIA A100 multi-instance GPU (MIG) 2g.20gb partition. Validation uses all streams, deterministic routing, and a repeated mask sequence with mask seed 44. The checkpoint minimizing total validation loss is selected without downstream labels.

Following the archive-pixel convention used by CSMoE, pretraining scale is $1{,}228{,}121\times64^2=5.030$G spatial pixels. This excludes channels, held-out samples, and repeated epochs, and is not a count of unique geographic ground area. Recorded elapsed time is approximately 166.9 h on the MIG partition; no conversion to full-GPU hours is assumed.

\subsection{Frozen Transfer and Spatial Protocols}
GEO-Bench v1 supplies official task splits and normalization statistics \citep{lacoste2023geobench}. All physically available bands compatible with the checkpoint are used (Table~\ref{tab:inputs}). The encoder is frozen with deterministic routing, and no training augmentation is applied to the downstream heads.

The 12-band tasks omit S2 B10. Physical band names, rather than array offsets alone, determine the model channel mapping. So2Sat does not supply orbit direction, so its VV/VH stream uses the ascending adapter once, without duplication into both SAR adapters. GEO-Bench BigEarthNet has 43 labels, distinct from the 19-label retrieval nomenclature. All four metadata groups are absent from these benchmark loaders.

Two protocols are fixed in advance: \emph{native-grid 64} resizes input imagery to the pretraining dimensions, whereas \emph{standard-grid 224} uses \(224\times224\) imagery. Invalid raster pixels do not contribute to nodata-aware bilinear input resizing. Classification labels are unchanged. For segmentation, both protocols retain \(224\times224\) ground-truth class maps and evaluate logits at that resolution; nearest-neighbor interpolation preserves class IDs when resizing labels, while model logits are resized bilinearly. Thus, a 64-pixel input does not imply 64-pixel supervision or evaluation. Resizing changes image sampling and token count, not the physical geographic extent of a sample, an important distinction because preprocessing itself affects EO benchmarks \citep{corley2024preprocessing}.

Classification averages the final pre-norm fine tokens into a 144-dimensional feature and trains one linear layer. AdamW uses zero weight decay for 50 epochs. The learning rates are $10^{-2}$ for BigEarthNet and So2Sat, $3\times10^{-3}$ for brick-kiln, and $5\times10^{-2}$ for EuroSAT\@. BigEarthNet uses GEO-Bench's balanced binary cross-entropy and selects the epoch with the highest validation micro-average precision; the single-label tasks use cross-entropy and validation average class accuracy. Results are means and population standard deviations over five independently initialized heads.

Segmentation uses the final pre-norm fine-token map. The same final map is supplied to all four inputs of a UPerNet head, which learns spatial scales of $4$, $2$, $1$, and $0.5$ relative to the token grid. The head has 256 decoder channels, BatchNorm/ReLU blocks, and pyramid-pooling scales $(1,2,3,6)$. Adam trains it for 50 epochs with cross-entropy, using learning rate $3\times10^{-4}$ for cashew and $10^{-2}$ for SA crop type. Validation mIoU selects the checkpoint for each of five head initializations.

For class $c$, recall is $\mathrm{TP}_c/(\mathrm{TP}_c+\mathrm{FN}_c)$ and intersection over union is $\mathrm{TP}_c/(\mathrm{TP}_c+\mathrm{FP}_c+\mathrm{FN}_c)$. Average class accuracy (AA) and mIoU are the unweighted means of these quantities over classes. Pixel accuracy is $\sum_c\mathrm{TP}_c$ divided by the number of evaluated pixels. For multilabel BigEarthNet, micro-averaged average precision (micro-mAP) pools all sample--class scores before computing average precision; macro-mAP averages classwise average precision, and micro-F1 pools true positives, false positives, and false negatives across sample--class decisions.

\begin{table}[!t]
\caption{Evaluation inputs and encoded metadata.}
\label{tab:inputs}
\centering\footnotesize
\begin{tabularx}{\columnwidth}{@{}lYl@{}}
\toprule
Dataset/task & Raster input & Lat./lon./month/ERA5 \\
\midrule
m-BigEarthNet & S2, 12 bands & All absent \\
m-brick-kiln & S2, 13 bands & All absent \\
m-EuroSAT & S2, 13 bands & All absent \\
m-So2Sat & S2, 10; S1 VV/VH & All absent \\
m-cashew-plant & S2, 12 bands & All absent \\
m-SA-crop-type & S2, 12 bands & All absent \\
BENv2-14k CBIR & S2, 10; S1 VV/VH & All absent \\
MMEarth WorldCover & S2 and/or S1 streams & Present/ablated \\
\bottomrule
\end{tabularx}
\end{table}

\subsection{Comparison and Analysis Controls}
CSMoE is the closest published comparison because it reports the same six tasks with frozen encoders, linear classification heads, and UPerNet segmentation heads \citep{hackel2026csmoe}. We use its exact Table IV values and identify other-model values digitized from its comparison figures as approximate. We do not rerun competing models. The comparison matches tasks, primary metrics, and the broad probing regime, not every input band, pretraining distribution, or implementation choice. CSMoE uses CLS features for classification; our pooling policy is selected independently on validation data.

Routing analysis uses 128 validation samples selected with seed 42 and the best checkpoint. Seven stream combinations and a masked-input pass distinguish stream availability from spatial masking. Returned sparse gates are accumulated by token count, with the three layer-0 streams reported separately from fused fine tokens; metadata and CLS are excluded from spatial statistics. If $p_t$ is the dense clean router distribution for token $t$, normalized local entropy is $T^{-1}\sum_t H(p_t)/\log E$, while normalized global entropy is $H(T^{-1}\sum_t p_t)/\log E$ for $E$ experts.

Absolute-position dependence uses normalized mutual information (NMI), with mutual information divided by the geometric mean of the position and expert entropies. Five patch-level permutations shuffle expert labels while preserving their global counts. The semantic diagnostic reduces valid Dynamic World pixels to one majority class per model patch and applies the same NMI calculation to class and top-1 expert assignments. To compare expert functions, every expert in a layer is evaluated on the same captured normalized tokens and pairwise output cosine similarity is averaged over those tokens. The intervention test uses eight fixed images and mask seed 44; one expert output is set to zero while its router assignments and all remaining computation stay unchanged. These measurements are checkpoint diagnostics on the stated samples, not archive-wide estimates.

\subsection{Metadata Availability Protocol}

We measure the effect of metadata without repeating pretraining. The 12,405 samples in the MMEarth pretraining-validation pool are divided into 7,443 probe-training, 2,481 validation, and 2,481 test samples. Samples whose source latitude and longitude match after rounding to five decimal places remain in the same partition. The encoder is frozen, and all raster inputs remain available throughout the experiment. Because the images were used for unsupervised checkpoint selection but their WorldCover labels were not used during pretraining, this is a held-out-label diagnostic on the pretraining corpus rather than an independent downstream benchmark.

A 1,595-parameter $1\times1$ convolutional head maps the final $16\times16$ pre-norm patch map to 11 WorldCover logits, which are bilinearly upsampled to the native $64\times64$ label grid. Invalid WorldCover pixels use ignore index $-100$ and do not contribute to cross-entropy or the confusion matrix. The head is trained with AdamW for up to 50 epochs using a learning rate of $10^{-3}$, weight decay of $10^{-4}$, batch size 128, and validation-mIoU selection with patience 10. We compare five metadata settings: all metadata, no metadata, no latitude or longitude, no month, and no ERA5.

Two protocols distinguish adaptation to missing metadata from unexpected removal at inference. In the \emph{matched} protocol, a separate head is trained and evaluated with each metadata setting. In the \emph{test-time removal} protocol, the head trained with all metadata is kept fixed while selected metadata groups are removed only during evaluation. Both protocols use the same frozen encoder, raster inputs, data partitions, and valid-pixel mIoU calculation.

\subsection{Controlled Sensor Availability and Degradation}
We evaluate input robustness without retraining either the encoder or downstream head. The experiment uses the same WorldCover probe partition as the metadata study. The all-stream, all-metadata $1\times1$ head is evaluated unchanged on the test set with seven clean stream subsets: S2; S1A; S1D; both S1 streams; S2+S1A; S2+S1D; and all streams.

S2 degradation selects exactly 25\%, 50\%, or 75\% of the 256 token locations, each corresponding to a $4\times4$-pixel region, and zeros the same locations in every S2 band. In the \emph{declared} condition, zeroed values are also marked invalid; in the \emph{undeclared} condition, the same values are zeroed while their validity indicators remain unchanged. The latter is a controlled interface error, not a claim that zero is a realistic cloud value. Four stream contexts---S2, S2+S1A, S2+S1D, and all streams---are tested with five corruption seeds. Metadata, normalization, routing mode, test samples, and head weights remain fixed. Standard deviations measure sensitivity to sampled mask locations only.

\subsection{Task Adaptation and Retrieval Protocols}
BigEarthNet adaptation keeps the same splits, normalization, band mapping, and pre-norm mean-fine pooling as the frozen evaluation. All encoder blocks and the linear head are optimized, while the 17,400 router parameters remain frozen. The encoder's maximum learning rate is $10^{-4}$ with layer-wise decay 0.75, and the head learning rate is $10^{-3}$. Training uses balanced binary cross-entropy plus the depth-averaged MoE balance loss at scale 1.0. One head-only epoch is followed by a two-epoch warmup and cosine decay over 20 epochs. Rotations and reflections are applied only to training images, and validation micro-mAP selects the checkpoint.

For content-based image retrieval (CBIR), BENv2-14k provides 13,683 paired optical/SAR images with 19 labels \citep{clasen2024reben}. The 3,255 validation images are queries and the 3,248 test images form the retrieval gallery. Inputs are resized to 64 pixels and normalized with MMEarth statistics. SAR channels are reordered from VH/VV to VV/VH, retained in decibels, and assigned to S1A or S1D from acquisition time. Exact cosine search retrieves $K=5$ neighbors from raw L2-normalized pre-norm mean-fine embeddings. For query labels $Y_q$ and retrieved-image labels $Y_r$, pair precision is $|Y_q\cap Y_r|/|Y_r|$ and pair recall is $|Y_q\cap Y_r|/|Y_q|$. Precision and recall are averaged over all query--neighbor pairs before their harmonic mean is computed, following CSMoE.

\section{Results}
\subsection{Learning and Frozen Transfer}
Validation reconstruction falls from 0.1598 at epoch 1 to 0.1046 at the selected epoch 31; total validation loss is 0.1146 and the weighted balance term remains near 0.0100. Epoch 50 reconstruction is 0.1057, confirming that later training does not improve checkpoint selection. These normalized losses are not directly comparable across target scalings.

\begin{figure}[!h]
\centering
\includegraphics[width=\textwidth]{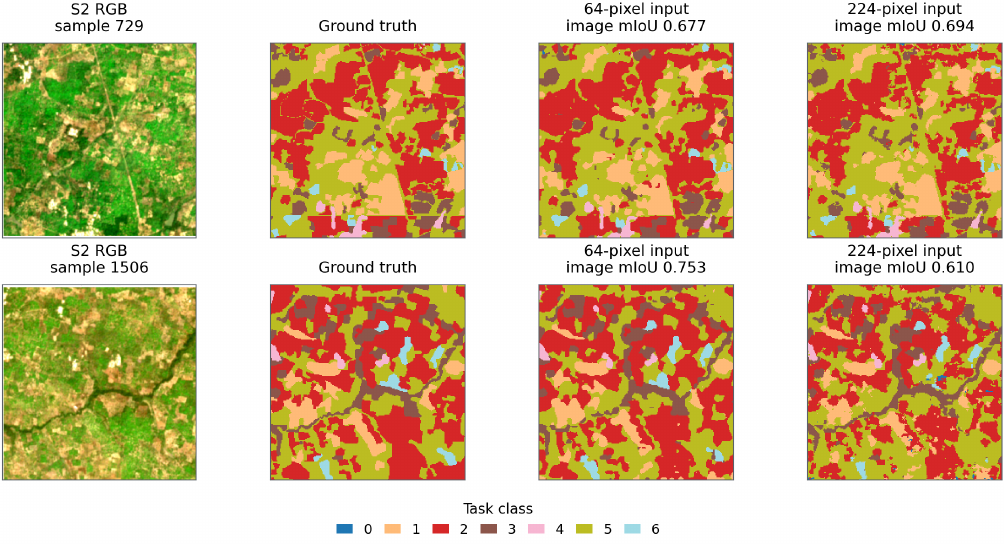}
\caption{Cashew segmentation for two fixed test scenes from the visualization notebook's seeded random sample. The same seven task-class colors are used for ground truth and predictions. Per-image mIoU is illustrative; aggregate results are in Table~\ref{tab:transfer}.}
\label{fig:cashew}
\end{figure}

Table~\ref{tab:transfer} reports frozen transfer. At 64 pixels, EuroSAT reaches 89.70\% AA and cashew 64.42\% mIoU, exceeding the strongest reported CSMoE variants by 1.40 and 5.02 points under their respective frozen protocols. Figure~\ref{fig:cashew} shows two cashew examples. The 224 grid improves four tasks, most notably crop type by 4.68 points, but changes cashew little and reduces So2Sat by 2.43 points. Spatial resampling is therefore task dependent. BigEarthNet and crop type remain the principal frozen-transfer gaps, and the strong cashew result should not be generalized to every agricultural task.
\begin{table}[!h]
\caption{Frozen GEO-Bench transfer (\%). Ours is mean $\pm$ standard deviation across five heads. Comparator values follow CSMoE \citep{hackel2026csmoe}; $\dagger$ marks figure-digitized values. Bold exceeds every CSMoE variant for that task.}
\label{tab:transfer}
\centering\footnotesize
\resizebox{\textwidth}{!}{%
\begin{tabular}{@{}lrrrrrrr@{}}
\toprule
Model & Input & BigEarthNet & Brick-kiln & EuroSAT & So2Sat & Cashew & SA crop type \\
 & & micro-mAP & AA & AA & AA & mIoU & mIoU \\
\midrule
CSMoE $P=32$ &224&62.6&93.2&84.9&44.1&46.0&35.8\\
CSMoE $P=28$ &224&65.1&93.8&84.9&46.3&48.3&36.7\\
CSMoE $P=16$ &224&66.5&94.3&86.2&48.0&55.7&38.6\\
CSMoE $P=14$ &224&66.0&94.4&88.3&49.6&59.4&39.8\\
DOFA$^\dagger$ &224&62.0&94.4&93.9&57.4&56.3&38.0\\
Prithvi V2-600$^\dagger$ &224&66.2&95.3&93.3&55.5&61.9&41.8\\
TerraMind$^\dagger$ &224&72.2&94.9&94.5&57.1&58.8&42.8\\
\midrule
\model &64&$55.16\pm0.16$&$91.60\pm0.54$&$\mathbf{89.70}\pm0.30$&$46.02\pm0.32$&$\mathbf{64.42}\pm0.22$&$27.65\pm0.47$\\
\model &224&$57.22\pm0.11$&$92.04\pm0.23$&$\mathbf{90.56}\pm0.27$&$43.59\pm0.76$&$\mathbf{64.08}\pm1.07$&$32.33\pm0.23$\\
\bottomrule
\end{tabular}}
\end{table}
\subsection{Parameter and Computation Trade-Off}
Table~\ref{tab:compute} separates full-model storage from encoder-only computation. Floating-point operations (FLOPs) count one multiply-add as one operation for batch-one S2 extraction and exclude the decoder, downstream head, and unsupported elementwise operations; they are not end-to-end UPerNet cost or throughput.

\begin{figure}[!h]
\centering
\includegraphics[width=\textwidth]{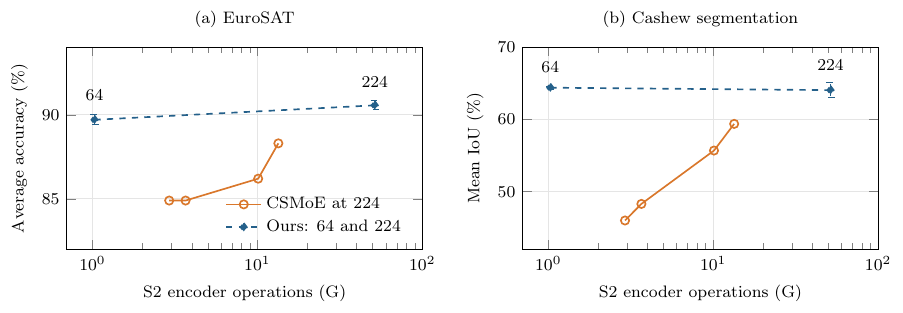}
\caption{Frozen transfer versus counted S2 encoder operations. The same \model checkpoint is evaluated at 64 and 224 pixels; CSMoE variants use 224 pixels. Axes are logarithmic, error bars show head-run standard deviations, and lines only connect related operating points.}
\label{fig:efficiency}
\end{figure}

The complete model has approximately $1/87$ as many parameters as CSMoE $P=14$ and uses $1/2.88$ as many pretraining archive pixels. Native-grid S2 inference uses $1/2.84$ as many counted operations as CSMoE $P=32$. At 224, however, $P=4$ gives 3,136 patches and global attention raises the count to 51.477G. The parameter advantage remains, but the compute advantage does not (Figure~\ref{fig:efficiency}).
\begin{table}[!h]
\caption{Model scale. Parameter rows report the complete pretraining model; the MEOX encoder alone has 2.939M parameters. FLOPs cover S2 encoder extraction, and pretraining pixels (PT PIX) count archive pixels.}
\label{tab:compute}
\centering\footnotesize
\begin{tabular}{@{}lrrrr@{}}
\toprule
Model & Input & Params (M) & FLOPs (G) & PT PIX (G) \\
\midrule
\model &64&3.115&1.029&5.030\\
\model &224&3.115&51.477&5.030\\
CSMoE $P=32$ &224&277&2.92&14.5\\
CSMoE $P=28$ &224&275&3.67&14.5\\
CSMoE $P=16$ &224&271&10.11&14.5\\
CSMoE $P=14$ &224&271&13.40&14.5\\
\bottomrule
\end{tabular}
\end{table}

\subsection{Routing and Expert Specialization}
Expert use is measured by execution rate, gate share, and top-1 rate. Execution rate is the fraction of tokens that select an expert among their top two and therefore sums to 200\% across experts. Gate share averages the sparse post-selection routing weight and sums to 100\%, while top-1 rate counts only the highest-ranked expert and also sums to 100\%. Across fused layers 1--14, the minima over all layer--expert pairs are 25.12\% execution rate, 11.55\% gate share, and 9.28\% top-1 rate. Every expert therefore receives tokens on the 128-sample diagnostic set. Normalized dense-router entropy is 0.984--0.999 after aggregating probabilities over tokens and 0.886--0.956 when entropy is computed per token and then averaged. The routers are balanced in aggregate while individual token distributions remain more selective.

\begin{figure}[!h]
\centering
\includegraphics[width=\textwidth]{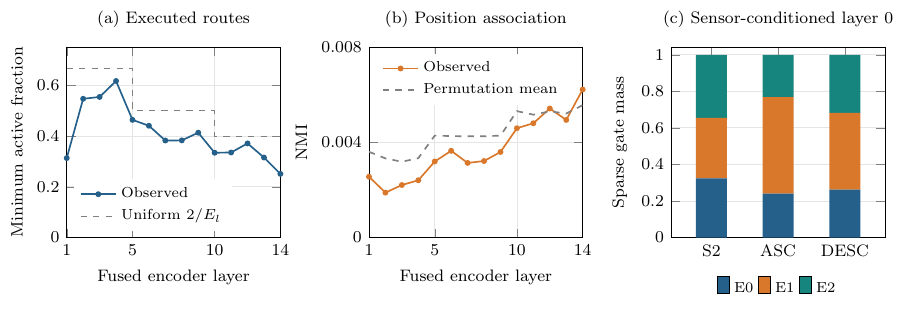}
\caption{Routing diagnostics for the selected checkpoint on 128 validation samples. Left: minimum fraction of fine tokens executing an expert, with the uniform top-2 fraction $2/E_l$ for context. Middle: absolute-position/top-1 NMI and its permutation mean. Right: layer-0 gate mass by stream. These measurements distinguish active dispatch, spatial association, and stream-conditioned routing; none alone establishes semantic specialization.}
\label{fig:routing}
\end{figure}

We next test whether routers repeatedly assign experts according to absolute image position rather than image content. For each fused layer, we compute the normalized mutual information (NMI) between patch position and the top-1 expert. We compare this value with shuffled expert assignments that preserve the overall number of tokens assigned to each expert. The observed NMI ranges from 0.00189 to 0.00623 and remains close to the shuffled baseline (Figure~\ref{fig:routing}). At layer 14, for example, the observed NMI is 0.00623 compared with a permutation mean of 0.00557. This small excess provides no evidence of a fixed spatial routing template, although attention can still transmit spatial information to the router.

Stream-dependent routing is visible in the shared pre-fusion block. Expert 1 receives an average gate share of 33.05\% for S2 tokens, 52.86\% for S1A tokens, and 42.02\% for S1D tokens. Because the same block and router process every stream, these differences show that routing responds to stream-conditioned representations. Expert 1 is not exclusively a SAR expert because it also processes S2 tokens. Separately, the fusion module assigns mean weights of 25.03\%, 37.83\%, and 37.14\% to S2, S1A, and S1D. In a hard invalidation check that marks an entire stream invalid at every position, its maximum fusion weight is approximately \(1.5\times10^{-6}\) when valid alternatives are present. The progressive experiment below invalidates selected S2 patches while retaining valid S2 context elsewhere and is therefore reported separately.

Dynamic World class/top-1-expert NMI ranges from 0.0270 to 0.1232, compared with permutation means of 0.00013--0.00031. Layer 2 has the largest observed association. This supports land-cover-dependent routing without requiring one expert per semantic class. Evaluating different experts on the same normalized tokens gives output cosine similarities of approximately 0.68--0.89 in early layers and 0.39--0.69 in deeper five-expert layers, showing differentiation despite shared projections.

Finally, we tested whether the experts selected by the router contribute to reconstruction. The same masked batch of eight images was processed repeatedly, and one expert output was set to zero in each run. The router selections were kept fixed, and the removed expert weight was not reassigned to another expert. Later layers therefore received the altered representation and continued processing it normally. Without suppression, the reconstruction loss was 0.08933. Removing any layer-0 expert increased the loss by between 0.00215 and 0.00334, showing that all three experts affected the reconstruction. The increases were smaller at layer 7, ranging from 0.00014 to 0.00062. At layer 14, removing expert 0 increased the loss by 0.00325, whereas removing another expert increased it by between 0.00013 and 0.00090. Expert contributions are therefore unequal for this batch, particularly in the final layer. This experiment measures the immediate effect of removing an expert from the trained model. It does not estimate performance after retraining the model without that expert, and results from eight images should not be interpreted as population averages.
\subsection{Embedding Selection and Geometry}
We use a predefined EuroSAT pilot containing 2,000 training and 1,000 validation samples to select the encoder representation. The selected representation is then fixed for all main experiments. Table~\ref{tab:embedding} compares tokens taken before and after the encoder's final LayerNorm and several pooling strategies. Averaging all pre-norm tokens gives the highest AA, 89.68\%, but averaging only the fine spatial tokens is within the predefined 0.5-percentage-point tie threshold. We therefore select pre-norm mean-fine features because they exclude CLS and metadata tokens and correspond directly to the spatial features used for dense prediction. Concatenating CLS and metadata tokens provides no improvement in this pilot.

We next examine whether the embedding space is dominated by a small number of directions. For centered covariance eigenvalues $\lambda_i$, let $p_i=\lambda_i/\sum_j\lambda_j$; the effective rank is
\begin{equation*}
r_{\mathrm{eff}}
=
\exp\left(-\sum_i p_i\log p_i\right).
\end{equation*}
On 1,000 EuroSAT validation images, the effective rank is 8.44 out of 144 for pre-norm features and 6.18 for post-norm features. Mean pairwise cosine similarity is 0.5406 for pre-norm features and 0.5347 for post-norm features. After subtracting the training-set mean, these values fall to 0.0104 and 0.0129, respectively. Thus, much of the high cosine similarity comes from a shared mean direction, while the low effective rank shows that the remaining variance is still concentrated in relatively few dimensions. Per-dimension standardization increases the effective ranks to 15.00 and 15.61 for pre-norm and post-norm features, but a more uniform geometry does not necessarily preserve task-relevant information \citep{roy2007effectiverank}.

We test this directly by centering the embeddings, removing the $k$ leading principal components, and applying L2 normalization, with all transformation statistics fitted only on training data. For pre-norm features, validation AA decreases from 89.0\% at $k=0$ to 88.8\% at $k=1$, 86.8\% at $k=2$, and 70.1\% at $k=8$. The dominant components therefore contain useful semantic information rather than only nuisance variation. Per-dimension standardization also remains below raw pre-norm features in the linear probe, so the main experiments retain raw features.

Finally, we evaluate input normalization independently of embedding post-processing. The EuroSAT pilot obtains 90.60\% validation AA with MMEarth statistics, 89.60\% with statistics estimated from the EuroSAT training split, 89.52\% with official GEO-Bench statistics, and 61.36\% without z-scoring. Although MMEarth normalization performs best in this pilot, all main GEO-Bench experiments use the official task statistics to preserve comparability with published results. These experiments separately evaluate token selection, embedding geometry, and input normalization rather than treating them as the same problem.
\begin{table}[!h]
\caption{EuroSAT Validation Representation Selection (AA, \%). ``All'' includes prefix tokens; ``fine'' includes spatial patches only. Selection uses validation, not the final test scores.}
\label{tab:embedding}
\centering\footnotesize
\begin{tabular}{@{}lrrr@{}}
\toprule
Pooling & Dim. & Pre-norm & Post-norm \\
\midrule
Mean all &144&$89.68\pm0.18$&$89.56\pm0.22$\\
Mean fine (selected) &144&$89.52\pm0.22$&$89.40\pm0.14$\\
CLS + mean fine &288&$86.84\pm0.73$&$89.00\pm0.20$\\
CLS + metadata &720&$81.02\pm0.56$&$81.70\pm0.83$\\
CLS &144&$80.66\pm0.36$&$78.40\pm0.41$\\
\bottomrule
\end{tabular}
\end{table}

\subsection{Metadata Availability on Held-Out WorldCover}
All metadata yields 30.45\% test mIoU (Table~\ref{tab:metadata}). Removing everything costs 0.64 points when the head is retrained without metadata and 0.72 points when the all-metadata head is evaluated after test-time removal. ERA5 contributes the largest isolated measured change; month removal has little effect. These effects are not additive because metadata groups and image content can be redundant. Nine of eleven classes improve with all metadata relative to none, and the matched all-metadata and no-metadata heads disagree on 3.46\% of valid pixels. The result supports a modest positive role for context and operation without metadata. It does not establish that metadata was unnecessary during pretraining or predict its benefit in every downstream archive.
\begin{table}[!h]
\caption{WorldCover metadata ablation: test mIoU (\%). Matched settings retrain the head; removal keeps the all-metadata head.}
\label{tab:metadata}
\centering\footnotesize
\begin{tabular}{@{}lrrrr@{}}
\toprule
Metadata & Matched & $\Delta$ & Removal & $\Delta$ \\
\midrule
All &30.454&0&30.454&0\\
None &29.814&$-0.640$&29.730&$-0.724$\\
No location &30.333&$-0.120$&30.284&$-0.170$\\
No month &30.450&$-0.003$&30.400&$-0.054$\\
No ERA5 &30.162&$-0.292$&30.139&$-0.315$\\
\bottomrule
\end{tabular}
\end{table}

\subsection{Sensor Availability and S2 Degradation}
Complete stream removal does not cause catastrophic failure of the fixed all-stream head (Table~\ref{tab:sensor-removal}). S2 alone obtains 30.53\% mIoU, while the two S1 streams obtain 24.79\%. S2+S1A has the highest mIoU, 31.82\%, but the all-stream condition has substantially higher pixel accuracy than either S2 alone or S2+S1A (61.54\% versus 55.41\% and 59.61\%). The apparent mIoU gain from removing S1D therefore reflects class-specific trade-offs rather than a universal benefit: S2+S1A improves several rare classes but reduces IoU for common water, tree, grassland, and cropland classes. The complete three-stream input also has the lowest cross-entropy, 1.10 versus 1.56 for S2 alone and 1.21 for S2+S1A.

\begin{table}[!h]
\caption{WorldCover performance of one fixed all-stream head under complete stream removal (\%). No condition retrains the encoder or head.}
\label{tab:sensor-removal}
\centering\footnotesize
\begin{tabular}{@{}lrrr@{}}
\toprule
Streams & mIoU & $\Delta$ mIoU & Pixel accuracy \\
\midrule
All &30.45&0.00&61.54\\
S2 &30.53&$+0.08$&55.41\\
S1A+S1D &24.79&$-5.66$&55.55\\
S1A &23.23&$-7.22$&51.69\\
S1D &22.46&$-7.99$&52.30\\
S2+S1A &31.82&$+1.37$&59.61\\
S2+S1D &30.24&$-0.21$&59.92\\
\bottomrule
\end{tabular}
\end{table}

\begin{figure}[!h]
\centering
\includegraphics[width=\textwidth]{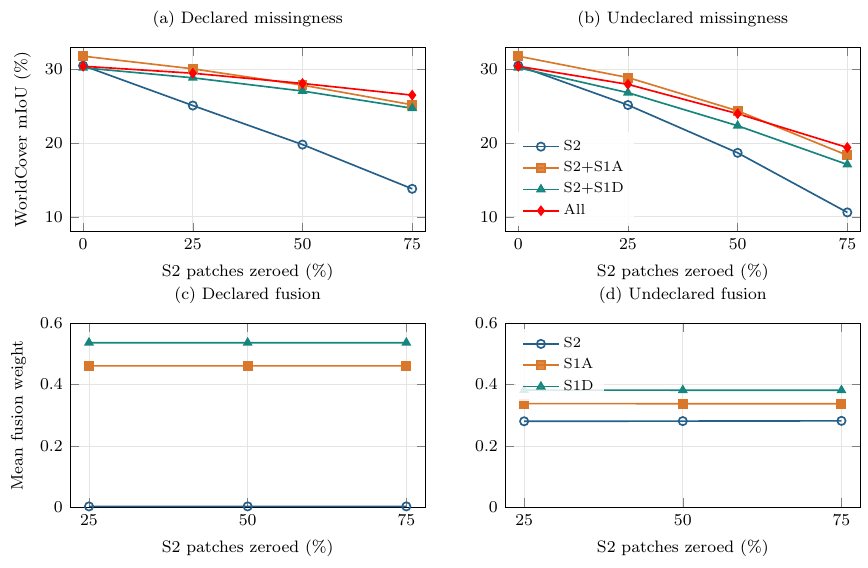}
\caption{Frozen WorldCover robustness. Top: mIoU under progressive S2 patch loss when zeroed patches are declared invalid or left undeclared. Curves show means over five corruption masks; variation is at most 0.09 mIoU points. Bottom: all-stream fusion weights measured only on zeroed S2 patches. Explicit validity redirects weight from S2 to the available SAR streams.}
\label{fig:sensor-robustness}
\end{figure}

Figure~\ref{fig:sensor-robustness} reports two controlled comparisons. The first tests whether S1 can compensate for missing S2 observations. When all streams are available and the removed S2 patches are marked invalid, mIoU is 29.50\%, 28.11\%, and 26.53\% after removing 25\%, 50\%, and 75\% of the S2 patches, respectively. The clean all-stream score is 30.45\%, so performance falls by 3.92 points at 75\% removal. With S2 as the only raster input, the same level of removal reduces mIoU from 30.53\% to 13.81\%. At 75\% removal, the all-stream input therefore outperforms S2 alone by 12.72 points. This difference indicates that the two S1 streams provide useful information where optical observations are missing.

The second comparison tests whether missing S2 patches must be identified through the validity mask. The raster values are identical in both conditions: selected S2 patches are replaced with zeros while both S1 streams remain available. When the zeroed patches are marked invalid, mIoU is 29.50\%, 28.11\%, and 26.53\% at 25\%, 50\%, and 75\% removal. If the same patches remain marked as valid, the corresponding scores are 27.99\%, 24.01\%, and 19.42\%. Providing the correct validity mask therefore improves mIoU by 1.51, 4.10, and 7.11 points as the amount of missing S2 data increases.

The fusion weights show how the validity mask changes the model response. For clean all-stream inputs, the average weights are 30.0\% for S2, 32.9\% for S1A, and 37.1\% for S1D. At locations where S2 is marked invalid, its average weight falls to approximately 0.3\%, while the S1A and S1D weights rise to approximately 46.1\% and 53.6\%. When the same zeroed S2 patches remain marked as valid, S2 continues to receive approximately 28.0--28.2\% of the fusion weight. The fusion module does not identify these zeroed patches as missing on its own. It redirects weight toward S1 when the validity mask reports that the S2 observation is unavailable, and the mIoU differences show that this redirection improves the segmentation result.

\subsection{Task Adaptation and Retrieval}
BigEarthNet micro-mAP rises from 55.16\% to 67.78\% at 64 pixels and from 57.22\% to 72.95\% at 224 pixels after encoder adaptation (Table~\ref{tab:finetune}). Macro-mAP and micro-F1 improve under both spatial protocols. These results are reported separately from frozen probing because the encoder weights, except for the routers, are updated with task labels.

On BENv2-14k, same-sensor F1 is 64.41\% for S1 and 66.33\% for S2 (Table~\ref{tab:retrieval}); cross-sensor F1 is lower. S1 retrieval exceeds the cross-sensor masked autoencoder (CSMAE) and the two coarser CSMoE variants, whereas S2 and the strongest cross-sensor directions remain below the best references. Multimodal reconstruction therefore yields useful within-sensor semantics but does not produce a fully aligned radar--optical metric space.
\begin{table}[!h]
\caption{BigEarthNet adaptation (\%). Finetuning includes spatial augmentation and is distinct from frozen probing.}
\label{tab:finetune}
\centering\footnotesize
\begin{tabular}{@{}llrrr@{}}
\toprule
Input & Encoder & Micro-mAP & Macro-mAP & Micro-F1 \\
\midrule
64 &Frozen&55.16&44.32&45.89\\
64 &Adapted&67.78&58.43&54.17\\
224 &Frozen&57.22&47.17&47.25\\
224 &Adapted&72.95&66.82&62.67\\
\bottomrule
\end{tabular}
\end{table}
\begin{table}[!t]
\caption{BENv2-14k retrieval at $K=5$ (\%). Comparator F1 values follow CSMoE Table VII \citep{hackel2026csmoe}; \model precision and recall appear below its F1.}
\label{tab:retrieval}
\centering\footnotesize
\begin{tabular}{@{}lrrrr@{}}
\toprule
Model/metric & S1$\to$S1 & S2$\to$S2 & S1$\to$S2 & S2$\to$S1 \\
\midrule
CSMAE &60.71&68.62&36.20&42.31\\
CSMoE $P=32$ &63.59&71.04&35.16&45.01\\
CSMoE $P=28$ &63.92&72.24&39.91&39.50\\
CSMoE $P=16$ &66.58&72.33&35.93&50.74\\
CSMoE $P=14$ &67.32&73.34&42.22&46.76\\
\midrule
\model F1 &64.41&66.33&39.49&40.00\\
\quad Precision &62.39&65.42&37.76&45.03\\
\quad Recall &66.57&67.26&41.39&35.98\\
\bottomrule
\end{tabular}
\end{table}
\section{Discussion and Conclusion}
The routing analysis indicates that the shared pre-fusion block develops stream-conditioned computation despite sharing its parameters across all three streams. In layer 0, expert 1 receives 33.05\% of the gate mass for S2 tokens, 52.86\% for S1A, and 42.02\% for S1D. All experts remain active, and suppressing any layer-0 expert increases masked reconstruction loss by 0.00215--0.00334. The association between absolute patch position and top-1 expert also remains close to the shuffled reference. These observations associate the routing differences with stream-conditioned representations rather than unused experts or fixed spatial assignments. The route-then-fuse design concentrates this processing in one shared pre-fusion block, allowing the remaining fourteen blocks to operate on \(N+5\) rather than \(3N+5\) tokens.

This organization yields a 2.939M-parameter encoder and a 3.115M-parameter pretraining model. Frozen transfer is strongest on EuroSAT and cashew segmentation, where MEOX reaches 90.56\% average class accuracy and 64.42\% mIoU, exceeding the strongest reported CSMoE variants by 2.26 and 5.02 points, respectively. Performance is less uniform on BigEarthNet and SA crop type. On BigEarthNet, full-encoder adaptation increases micro-average precision from 57.22\% to 72.95\%, indicating that the compact pretrained representation remains adaptable when linear transfer is insufficient. The model is computationally efficient at its native 64-pixel resolution; at 224 pixels, however, global attention over 3,136 patches increases encoder cost to 51.477G counted operations. The larger-grid experiments should therefore be interpreted as evidence of spatial flexibility rather than low-cost inference.

The WorldCover experiment shows how validity-aware fusion uses complementary streams under incomplete observations. Removing 75\% of S2 patches reduces mIoU from 30.53\% to 13.81\% when S2 is the only raster input, whereas retaining both S1 streams limits the reduction from 30.45\% to 26.53\%. For identical zeroed S2 patches, marking them invalid yields 26.53\% mIoU compared with 19.42\% when they remain marked as valid. The mean S2 fusion weight at those locations falls from approximately 28\% to 0.3\%, with the weight redistributed to S1A and S1D. The validity signal therefore redirects the fused representation toward available radar observations.

These robustness results are limited to random patch removal within the MMEarth validation corpus and do not establish robustness to realistic cloud fields, radiometric degradation, unseen sensors, or arbitrary missing bands. Frozen crop-type segmentation and cross-sensor retrieval also remain weaker than the best published references. The current experiments do not include matched early-fusion or dense-encoder baselines, so they characterize the route-then-fuse operating point rather than isolating its causal advantage over every alternative. MEOX implements stream-conditioned sparse processing in one shared pre-fusion block followed by a common deep representation. This organization produces a 2.939M-parameter multimodal encoder and preserves useful radar information when optical patches are unavailable.

\section*{Code and Data Availability}
The implementation and experiment notebooks are maintained in the \href{https://github.com/AlbughdadiM/compact-multimodal-moe-eo}{public project repository}. Pretraining uses MMEarth64 and downstream evaluation uses GEO-Bench v1 and BENv2-14k. Dataset access remains subject to the respective providers' terms.

\section*{Declaration of Generative AI and AI-assisted Technologies in the Manuscript Preparation Process}
During the preparation of this work, the author used OpenAI Codex to assist with language editing. The author reviewed and edited all generated material and takes full responsibility for the content of the article.

\bibliographystyle{elsarticle-harv}
\bibliography{references}
\end{document}